%% file: naaclhlt2019.tex
\documentclass[11pt,a4paper]{article}
\usepackage[hyperref]{acl2021}
\usepackage{times}
\usepackage{latexsym}
\usepackage{url}
\usepackage{multirow}
\usepackage[ruled,vlined]{algorithm2e}
\usepackage{algorithmic}
\usepackage[utf8]{inputenc} % allow utf-8 input
\usepackage[T1]{fontenc} % use 8-bit T1 fonts
\usepackage{hyperref} % hyperlinks
\usepackage{url} % simple URL typesetting
\usepackage{booktabs} % professional-quality tables
\usepackage{amsfonts} % blackboard math symbols
\usepackage{nicefrac} % compact symbols for 1/2, etc.
\usepackage{microtype} % microtypography
\usepackage[utf8]{inputenc} % allow utf-8 input
\usepackage[T1]{fontenc} % use 8-bit T1 fonts
\usepackage{hyperref} % hyperlinks
\usepackage{url} % simple URL typesetting
\usepackage{booktabs} % professional-quality tables
\usepackage{amsfonts} % blackboard math symbols
\usepackage{nicefrac} % compact symbols for 1/2, etc.
\usepackage{microtype} % microtypography
\usepackage{graphicx}
\usepackage{xcolor}
\usepackage{gensymb}
\usepackage{subcaption}
\usepackage{amsthm}
\usepackage{mathrsfs}
\usepackage{amsmath}
\usepackage{amssymb}
\usepackage{mathtools}
\usepackage{wrapfig}
\usepackage{soul}
\makeatletter
\def\BState{\State\hskip-\ALG@thistlm}
\makeatother

\usepackage{stackengine}

\makeatletter
\newcommand{\distas}[1]{\mathbin{\overset{#1}{\kern\z@\sim}}}%

\newcommand{\context}{ \phi(q_i,c'_i)}
\newcommand{\question}{ \phi(q'_i,c_i)}

\newcommand{\beqs}{\vspace{0mm}\begin{eqnarray}}
\newcommand{\eeqs}{\vspace{0mm}\end{eqnarray}}
\newcommand{\barr}{\begin{array}}
\newcommand{\earr}{\end{array}}

\usepackage{bbm}

\aclfinalcopy % Uncomment this line for the final submission
\title{Knowing More About Questions Can Help: \\Improving Calibration in Question Answering}

\author{Shujian Zhang  \qquad Chengyue Gong \qquad  Eunsol Choi \\
The University of Texas at Austin \\
\texttt{szhang19@utexas.edu, \{cygong,eunsol\}@cs.utexas.edu}
}

\date{}

\begin{document}
\maketitle

\begin{abstract} We study calibration in question answering, estimating whether model correctly predicts answer for each question. Unlike prior work which mainly rely on the model's confidence score, our calibrator incorporates information about the input example (e.g., question and the evidence context). Together with data augmentation via back translation, our simple approach achieves 5-10\% gains in calibration accuracy on reading comprehension benchmarks. Furthermore, we present the first calibration study in the open retrieval setting, comparing the calibration accuracy of retrieval-based span prediction models and answer generation models. Here again, our approach shows consistent gains over calibrators relying on the model confidence. Our simple and efficient calibrator can be easily adapted to many tasks and model architectures, showing robust gains in all settings.\footnote{Code is available at \url{https://github.com/szhang42/Calibration_qa}.} 
\end{abstract}
\input{introduction}

\input{task-2}
\input{model-3}

\input{experiment-4}

\input{results-5}
\input{analysis-6}
\input{related-7}

\input{conclusion}

\section*{Acknowledgments}
We would like to thank UT Austin NLP group, especially Kaj Bostrom and Greg Durrett for feedback and suggestions. 
\bibliography{naaclhlt2019}
\bibliographystyle{acl_natbib}
\clearpage
\onecolumn
\input{appendix}

\end{document}

%% file: introduction.tex
\section{Introduction}
Despite rapid progress in AI models, building a question answering (QA) system that can always correctly answer any given query is beyond our reach. Thus, questioners have to {interpret} the model prediction, deciding whether to trust it. We study providing an accurate estimate of the correctness of model prediction for each example at test time. As making incorrect predictions can be much more costly than making no prediction (e.g., missing diagnosis is much more costly than querying human experts), calibrators can bring practical benefits~\cite{kamath2020selective}. 

Existing work on calibration focuses on mdoel confidence, such as the max probability of the predicted class~\cite{guo2017calibration,Desai2020CalibrationOP}. Unlike classification tasks, question answering explores large output space, either through answer generation~\cite{Raffel2020ExploringTL,Lewis2020RetrievalAugmentedGF} or selecting a span from provided documents~\cite{rajpurkar2016squad}. In both settings, optimal decoding is often prohibitively expensive, and heuristic decoding is a standard practice~\cite{Seo2017BidirectionalAF}. Thus, relying on the model's confidence score alone is not sufficient for calibration~\cite{Kumar2019CalibrationOE}. 

Nonetheless, prior work~\cite{kamath2020selective,Jagannatha2020CalibratingSO} relied heavily on model confidence, such as the max probability of the predicted answer, together with a handful of manually crafted features containing little information about the input, such as the length of the question. We empower the calibrator by introducing an input example embedding from a pre-trained language model~\cite{alberti2019bert,liu2019roberta} finetuned on QA supervision data as additional features. With this simple and general feature, calibrator can identify questions regarding rare entities or examples with little lexical overlap between the question and the context. We bring further gains by paraphrasing questions or contexts respectively through back translation~\cite{Sennrich2016ImprovingNM}, providing lexical variations of the question and the context and enriching the feature space.% the ambiguity of question.% to paraphrase questions and contexts. 

%\citep[e.g.][]{kamath2020selective, Desai2020CalibrationOP}. 
We evaluate our calibrator with internal metrics (i.e., calibration accuracy) and external metrics (i.e., impact on QA performance). We first evaluate calibrators in reading comprehension settings introduced in~\citet{kamath2020selective} -- in-domain~\cite{rajpurkar2016squad,kwiatkowski2019natural}, out of domain~\cite{fisch2019mrqa}, and adversarial~\cite{jia2017adversarial}. Then, we expand calibration study to more challenging open retrieval QA setting~\cite{trec, chen2017}, where a system is not provided with an evidence document. We adapt our calibrator for state-of-the-art generation based~\cite{Raffel2020ExploringTL} and extractive (retrieve-and-predict) QA models~\cite{karpukhin2020dense}, showing gains in both models. While calibration accuracy is higher in the generation based model, the extractive method provides better answer coverage above fixed accuracy. Lastly, we use calibrator as a reranker for the answer span candidates in an extractive open retrieval QA model~\cite{karpukhin2020dense}, showing modest gains. We provide rich ablation studies on design choices for our calibrator, such as the choice of base model to derive input example encoding. Our simple input example embedding from pretrained language models shows consistent gains in all settings and datasets. Without any manual engineering specific to the question answering task, our calibrator could be easily adapted to other tasks with rich output space. 
% To summarize, our contributions are the following:
% \begin{itemize}
%     \item Improving calibration accuracy 5-10\% by introducing rich contextual features through extracting question-context encodings from a pre-training language model and paraphrasing questions and contexts through back translation.
%     \item Comprehensive study and analysis of answer calibration in reading comprehension tasks, including previously-studied domain shift and new adversarial settings.
%     \item Proposing an efficient and general post-hoc calibrator which could be easily incorporated or extended to many other NLP tasks
%     \item Presenting the first study of calibration in open domain QA with both closed book and open book approaches, and applying calibrator for answer candidate reranking. 
% \end{itemize}

%% file: task-2.tex
\section{Problem Definition}\label{sec:task}
We estimate how the models' prediction confidence aligns with the empirical likelihood of correctness~\cite{Brier1950VERIFICATIONOF}. Formally, a calibrator $f$ takes the input example $x_i$ and the trained model $M_\theta$ and identifies whether the model's prediction is correct or not. We treat the correctness as binary (i.e., answer string exact match) for simplicity, instead of partial credit (e.g., token level F1 score). We study two settings: reading comprehension (RC) and open retrieval QA. In RC, an input example $x_i$ will be a context $c_i$ and the question $q_i$, and in open domain QA, an input example will be a corpus $C$ and the question $q_i$.

We use the same metrics to evaluate the performance of the calibrator $f$ in the two settings. 
\subsection{Metric: Calibrator performance}
\paragraph{Accuracy:} 
Given evaluation data $D_{\text{eval}}=\{(x_1,y_1),(x_2,y_2)\ldots(x_N,y_N)\}$ and a learned model $M_{\theta}$, we define the accuracy of the calibrator $f$ as:
$$
\text{acc}(f)=\sum_{i=1}^{N}\mathbbm{I}\bigg\{f(x_i, M_\theta)=\mathbbm{I} [M_\theta(x_i)=y_i]\bigg\}.
$$

\paragraph{AUROC: } Based on the above definition of the accuracy of the calibrator $f$, we computes the coverage -- fraction of evaluation data $D_{eval}$ that model makes prediction on -- and risk, the error at that coverage. We plot risk versus coverage graph, and measure the area under the curve, i.e., AUROC (Area Under the Receiver Operating Characteristic Curve)~\cite{hanley1982auroc}. 

% \paragraph{End task performance}

\subsection{Metric: End task performance}
We measure how the calibrator performance impacts QA performances. First, we study selective QA setting -- where we use calibrator score to decide which examples from $D_{eval}$ to make predictions. %% and study it as answer reranker, changing model's answer prediction. 

For the extractive model for open retrieval QA~\cite{karpukhin2020dense}, where multiple answer candidates are given, we further evaluate the performance of calibrator as a reranker and measure the answer span exact match (EM) score. 

\paragraph{Selective QA (coverage at fixed accuracy): }
We use the calibrator score to rank the \textit{examples} in the evaluation data. Specifically, we use the calibrator's confidence for the top answer candidate instead of model score to decide which examples in $D_{eval}$ the model answers most confidently. Then, we report the percentage of evaluation data that can be predicted while maintaining threshold accuracy (80\%), following prior work~\cite{kamath2020selective}.

\paragraph{Open Retrieval QA (top-N accuracy):}
We use the calibrator score to rank the \textit{answer candidates} for each evaluation example, similar to how candidate translations are reranked in machine translation~\cite{Shen2004DiscriminativeRF}. We first retrieve answer candidates from multiple paragraphs and utilize the calibrator to override the model's prediction. The calibrator scores the top N answer candidates and outputs the answer with the highest confidence score instead of the answer with the highest model score. Our calibrator can be added as last step for any open retrieval QA systems which generates multiple answer candidates without retraining the model. We evaluate the top 1 exact match accuracy and the top 5 exact match accuracy after re ranking with our calibrator score. 

%% file: model-3.tex
\section{Methods}
We propose two general approaches to improve binary calibrator: new feature vector, a dense representation of the input example (Section~\ref{subsec:feature}) and data augmentation with backtranslation which further improves the new feature vector (Section~\ref{subsec:paraphrase}). While both are simple, well-established formula for improving end tasks in NLP, neither has been explored in the context of calibration, as prior work assumed model confidence score is the most prominent signal. We follow prior work~\cite{kamath2020selective} for calibrator architecture and focus on improving its feature space. 

\subsection{Calibrator Architecture}
A binary classifier is trained using the gradient boosting library XGBoost~\cite{chen2016xgboost}, which classifies each test example as correctly answered by the base QA model or not. This calibrator does not share its weights with the base QA models. We finetune the following hyper-parameters on the development set: colsample by level, colsample by node, colsample by tree, learning rate, and the number of estimators. All calibrators are trained five times, each with different data partitions and random seeds. We report the variances in the results. 

\subsection{Input Example Embedding Feature From Base QA Model}~\label{subsec:feature}
Prior work uses manually designed features based on the scores to the predicted answer (details in Section~\ref{subsec:rc}). Such features retain little information about input example -- e.g, question and the evidence context. Inspired by the recent works in machine learning \citep[e.g.][]{song2019unsupervised, hendrycks2019using}, which use hidden vectors to classify in-domain and out-of-domain data, we introduce an input example embedding, a new feature vector that represent question and (optionally) evidence context to a calibrator. 

Our input example embedding is a fixed dimensional vector representing an input example, similar to sentence embeddings~\cite{Conneau2017SupervisedLO}. It differs in that the representation is taken from the final layer of base QA model, which is trained with supervision from question answering data and it encodes question and (optional) evidence context simultaneously. In Section~\ref{subsec:task-agnostic}, we report minor performance degradation from using embeddings from generic pretrained language model instead. 
%and training objective instead of natural language inference data and training objective.
%Our question context embeddings are similar to sentence embeddings~\cite{Conneau2017SupervisedLO}. But, our representations  

Each base model processes input example, either query $q_i$ or query, context pair $(q_i, c_i)$ to generate a sequence of hidden vectors, which will be compressed into a fixed dimensional vector to be used as calibrator feature.\footnote{For simplicity, we write equations with $(q_i, c_i)$ pair as an input, when only query is provided (e.g., generation based open retrieval QA method) $c_i$ is empty.} %We take the final hidden layer of a base QA model and average the final layer hidden representation across all time steps to get a vector for each input example.
We denote the input example as a sequence of tokens $\mathbf{t} = (t_0, t_1, \cdots, t_n)$ where $n$ is the length of the input. We pass the sequence $\mathbf{t}$ through base QA model and get $(\mathbf{h}_0, \mathbf{h}_1, \cdots, \mathbf{h}_n)$ where $\mathbf{h}_i$ is the corresponding final-layer hidden state of $t_i$, and $\mathbf{h}_i = (\mathbf{h}_{i, 0}, \cdots, \mathbf{h}_{i, m})$ where $m$ is the number of hidden dimensions.
Then, we get the $m$-dimensional feature vector 
\begin{equation}
    \label{eq:feature}
    \phi(q_i,c_i) = [\frac{1}{n}\sum_{i=1}^{n} \mathbf{h}_{i, 0}; \cdots;  \frac{1}{n}\sum_{i=1}^{n} \mathbf{h}_{i, m})],
\end{equation}
where each dimension is an average across the length $n$. 
We then train a binary classifier using these features as a calibrator. We now describe our base QA models to get this hidden representations.

%We describe the base QA model architectures and data used to train them. 

\subsubsection{Base QA Model}
We use standard span prediction architecture for RC, and a generation based model and an extractive model for open retrieval QA.

For RC and extractive open retrieval QA model, we use a standard span prediction architecture based on a pretrained language model~\cite{Devlin2018BERTPO}, which predicts start and end index of the answer span separately with softmax layer. The output hidden vector sequence will equal the sum of the length of question and the length of evidence context. For the open retrieval QA setting, the extractive model first retrieves a passage from the corpus and predicts an answer span from it. We use the best model from dense passage retrieval (DPR)\citep{karpukhin2020dense}.\footnote{\url{https://github.com/facebookresearch/DPR}} Specifically, this model retrieves the top $100$ retrieved passages as input and trains a span prediction model, which optimizes a softmax cross-entropy loss to select the correct passage among the candidates, and the answer span prediction loss. The model then selects the answer span with the highest answer span score (sum of the start and end logit score) from the passage with the highest passage score. 
In this setting, $\mathbf{t}$ is a concatenation of question $q_i$ and the context $c_i$. 

For generation based model, we use a sequence-to-sequence (seq2seq) model, specifically T5-small~\citep{Raffel2020ExploringTL}, which takes the question as an input and generates answer tokens. For this base QA model, $\mathbf{t}$ only consists of the query since the context is not provided.

\noindent \textbf{Data} For all experiments in RC, we train the model on the SQuAD 1.1 dataset. For open retrieval QA, models are trained on the Natural Questions (NQ) dataset ~\citep{kwiatkowski2019natural} following the data split from ~\citet{Lee2019LatentRF}.

\begin{table*}
\centering

\footnotesize

 \begin{tabular}{l|l|c|c|c}
 \toprule
 \bf{Task} & \bf{Setup}&  \bf{Base QA model (train)} & \bf{Calibrator (train \& dev)}  & \bf{Calibrator (test)} \\ \midrule
  & In domain&  & SQuAD 1.1 + HotpotQA  & SQuAD 1.1 + HotpotQA \\
   \multirow{2}{*}{Standard RC}  & Out of domain &  & SQuAD 1.1 + HotpotQA & Other MRQA  \\
 & In domain&  & SQuAD 1.1 + NQ & SQuAD 1.1 + NQ \\

  & Out of domain & SQuAD 1.1  & SQuAD 1.1 + NQ & Other MRQA  \\
 \multirow{2}{*}{Adversarial RC}& In domain & & SQuAD 1.1 Adversarial &  SQuAD1.1 Adversarial \\
 & Out of domain  &  & MRQA &  SQuAD1.1 Adversarial \\
 \multirow{2}{*}{Unanswerable RC}& In domain & & SQuAD 2.0  & SQuAD 2.0  \\
& Out of domain  & &MRQA  & SQuAD 2.0  \\
\hline \hline \\ [-2.0ex]
 Open Retrieval QA &In domain  & \multicolumn{2}{c|}{NQ Training Set} & NQ Test Set \\
 \bottomrule
  \end{tabular}
 \caption{Experiment Configuration. In domain / Out of domain distinguishes whether the training data for calibrator is different from the test data. } 
    \label{tab:experiement_setting}
\end{table*}

% Both models are trained on the train split~\cite{Lee2019LatentRF} of Natural Questions (NQ) dataset ~\citep{kwiatkowski2019natural}.

% We take the final hidden layer of a base QA model and average the final layer hidden representation across all time steps to get a vector for each input example. We denote the input example as a sequence of tokens $\mathbf{t} = (t_0, t_1, \cdots, t_n)$ where $n$ is the length of the input. For a sequence-to-sequence (seq2seq) generation model in open domain QA setting, $\mathbf{t}$ only consists of the query since the context is not provided. For other models, $\mathbf{t}$ is a concatenation of question $q_i$ and the context $c_i$.

% We pass the sequence $\mathbf{t}$ through base QA model and get $(\mathbf{h}_0, \mathbf{h}_1, \cdots, \mathbf{h}_n)$ where $\mathbf{h}_i$ is the corresponding final-layer hidden state of $t_i$, and $\mathbf{h}_i = (\mathbf{h}_{i, 0}, \cdots, \mathbf{h}_{i, m})$ where $m$ is the number of hidden dimensions. 

\subsection{Data Augmentation Via Paraphrasing}~\label{subsec:paraphrase}
Paraphrase generation can improve QA models~\cite{Yu2018QANetCL} by handling language variation. Compared to sentence retrieval \cite{du2020self} and language model based example generation \cite{anaby2020not}, backtranslation can capture the ambiguity of questions and answer\cite{singh2019xlda}. 
Given a $(q_i, c_i)$ pair, we use back translation~\cite{Sennrich2016ImprovingNM} to generate paraphrases of the question $q'_i$ from $q_i$ and the evidence context $c'_i$ from $c_i$.

We use standard transformer-based neural machine translation models~\cite{mariannmt} trained on WMT dataset.\footnote{\url{https://huggingface.co/transformers/model_doc/marian.html}} 
We first translate the original sentences to a pivot language and then translate them back to the source language. To guarantee translation quality, French and German are used as the pivot languages. We use beam search decoding with beam size as 4 and truncate the context length to 512, as the reading comprehension model truncates the context anyway. We analyze the quality of backtranslation in Section~\ref{subsec:qual_backtranslation}. %In practice, this back translation shows good quality as it only changes a small portion of the original answers.

We denote $(q'_i, c_i)$ as $\mathbf{t}^q = (t^q_0, \cdots, t^q_{n_q})$ and $(q_i, c'_i)$ as $\mathbf{t}^c = (t^c_0, \cdots, t^c_{n_c})$.
Here, $n_q$ and $n_c$ denote the length after backtranslating the question and context, respectively.
For $\mathbf{t}^q$ and $\mathbf{t}^c$, we pass 
them through the base QA model, get $\mathbf{h}^q$ and $\mathbf{h}^c$, and
extract the $m$-dimensional feature vector as in Eqn \eqref{eq:feature}, 
\begin{align}
\begin{split}
\label{eq:bt}
   \phi(q'_i,c_i) & =  [\frac{1}{n_q}\sum_{i=1}^{n_q} \mathbf{h}^q_{i, 0}, \cdots,  \frac{1}{n_q}\sum_{i=1}^{n_q} \mathbf{h}^q_{i, m}], \\
   \phi(q_i,c'_i)& =  [\frac{1}{n_c}\sum_{i=1}^{n_c} \mathbf{h}^c_{i, 0}, \cdots,  \frac{1}{n_c}\sum_{i=1}^{n_c} \mathbf{h}^c_{i, m})].
\end{split}
\end{align}

We use the concatenation of the original input example embedding and backtranslated one, $[\phi(q_i,c_i);\phi(q'_i,c_i)]$ and $[\phi(q_i,c_i);\phi(q_i,c'_i)]$ as features. Backtranslating both context and question did not bring further gains, thus the results from such a feature set are not presented. {We hypothesize that backtranslating context and question together might introduce too severe noise.} We do not use data augmentation for open retrieval QA experiments.

%% file: experiment-4.tex
\section{Experimental Settings}
In this section, we describe the experimental setting, dataset setups and baseline systems. Table~\ref{tab:experiement_setting} summarizes the evaluation scheme. A separate calibrator is trained for each calibrator train data configuration. %We test our calibrator under domain shift, in adversarial settings, and in open retrieval QA. %The detailed training details are included in the Appendix.  

\subsection{Data}
For all in-domain reading comprehension experiments, we randomly split the data into training, development, and test (40\%,10\%,50\%), following regression and classification benchmarks ~\cite{asuncion2007uci}. Further, we assume only limited supervised data is available for calibrators, simulating a set up where we have a general QA model and small number of annotated data reserved for calibration. 

\paragraph{Standard RC } We test two in domain settings and two out of domain settings. 
We randomly sample 4K examples from each of the datasets included in the training portion of the MRQA shared task~\cite{fisch2019mrqa} (SQuAD~\cite{rajpurkar2016squad}, NewsQA~\cite{trischler2017newsqa}, TriviaQA~\cite{joshi2017triviaqa}, SearchQA~\cite{dunn2017searchqa}, HotpotQA~\cite{yang2018hotpotqa}, Natural Questions~\cite{kwiatkowski2019natural}). We train two calibrators, one with the SQuAD1.1 + HotpotQA datasets and another with the SQuAD1.1 + NQ datasets. For out of domain evaluation, we use four remaining datasets from MRQA shared task training set.

\paragraph{Adversarial RC (SQuAD 1.1 Adversarial)} 
The adversarial examples manipulate the evidence paragraph to change the model prediction but not the gold answer. We sample 2K examples from the development portion of the SQuAD 1.1 ~\cite{jia2017adversarial} AddSent dataset, which appends an additional sentence that looks similar to the question at the end of the paragraph.  For the out-of-domain case, we train the calibrator on 6K examples (1K each sampled from MRQA datasets) and test on adversarial examples.

\paragraph{Unanswerable RC (SQuAD 2.0)} We sampled 2K examples from the development portion of the SQuAD 2.0 dataset~\cite{rajpurkar2018know}, which contains examples where the answer to the question cannot be derived from the provided context. Crowdworkers posed questions that were impossible to answer based on the paragraph alone while referencing entities in the paragraph and ensuring that a plausible answer is present. For out of domain setting, we train the calibrator on 6K examples (1K each sampled from MRQA datasets) and test on SQuAD 2.0 dataset (same as adversarial RC setting).

\paragraph{Open Retrieval QA}
We use the open retrieval version of the NQ ~\cite{Lee2019LatentRF}. We split its training data 60\% and 40\% for calibrator training and validation and use the NQ test set for testing.

\subsection{Comparison Systems}
We summarize the calibrators used in our study in Table~\ref{tab:mode_detail}. All calibrators are trained with the same gradient boosting library XGBoost~\cite{chen2016xgboost}, and they only differ in the feature sets. These calibrators are efficient, trained within a few minutes even with our new feature space.

\begin{table}
\centering
 \footnotesize \small
 \begin{tabular}{l|l|l}
 \toprule
 \bf{QA model} & \bf{Calibrator Feature Set} & \bf{\# Features} \\ %& \# Data \\
 \midrule
 \multirow{6}{*}{RC} &MaxProb & 1 \\ %& N \\
& features: \citet{kamath2020selective}  &17 \\ %& N\\
& Ours & $m$ \\
&  + features&  17 + $m$ \\ %& N\\
&  + features + $\context$ &17 + 2$m$\\ %& N \\
&  + features + $\question$ &17 + 2$m$ \\%& N\\
 \midrule
\multirow{2}{*}{Extractive} & 
% \citet{kamath2020selective} & 2 \\
 Unnormalized Scores & 2 \\
 & Normalized Scores & 2 \\
 (DPR) & Ours + Normalized Score & 2 + 2$m$ \\
 \midrule
Generation& Likelihood & 1\\ %& N \\
(T5)& Ours + Likelihood & 1 + $m$ \\
 
 \bottomrule
  \end{tabular}
 \caption{Comparison Systems: different calibrators explored for three base QA models. The last two QA models are for open retrieval QA task. The dimension of question context embedding is $m$ defined in Eqn \eqref{eq:feature} (eg. $m$ is 768 for reading comprehension).
 } 
    \label{tab:mode_detail}
\end{table}

\subsubsection{Reading Comprehension} \label{subsec:rc}
\paragraph{MaxProb} is the simplest baseline that relies on the model's confidence score. The model score is the sum of the logit scores of the start and end of the answer span
% of the start token of the answer span and the end token of the answer span logit scores
for reading comprehension. For open retrieval question answering, the model first determines the passage with the highest passage-match score and then extracts the answer span from this passage.

Formally, given the set of answer spans $Y$, MaxProb with
model $M_\theta$ estimates confidence on input $x_i$ as:
$$
\textit {MaxProb } =\max_{y \in Y} M_\theta \left(y \mid x_i \right),
$$ where $M_\theta (y \mid x_i )$ refers to the model score for candidate answer $y$.

\paragraph{\citet{kamath2020selective}} uses a calibrator based on the following general features: passage length, the predicted answer length, and the top-5 largest softmax probabilities generated by the model. They also use test time dropout~\cite{gal2016dropout}: given an input $x_i$ and model $M_\theta$, compute $M_\theta(x_i)$ with $K$ different dropout masks, obtaining prediction distributions $\hat p_1, ..., \hat p_k$, where each $\hat p_i$ is a probability distribution over $Y$.
Two options are used as confidence estimates. First, taking the mean of $\hat p_i$ \citep{lakshminarayanan2017simple}
$$
\textit {Dropout Mean }=\frac{1}{K} \sum_{i=1}^{K} \hat {p_i}.
$$
Second, taking the variance of the
$\hat p_i$ \cite{feinman2017detecting, smith2018understanding}
$$
\textit{Dropout Variance} = \operatorname{Var}\left[\hat{p}_{1}, \ldots, \hat{p}_{K}\right].
$$

\noindent The dimension of $\textit{MaxProb}$, 2th-5th probability, $\textit{Dropout Mean}$, $\textit{Dropout Variance}$, context length and prediction length are 1, 4, 5, 5, 1, 1, respectively. In total, this feature set contains 17 features. 

\begin{table*}
 \footnotesize
\centering
 \begin{tabular}{lccc|ccc}
 \toprule
  & \multicolumn{3}{c|}{In Domain} & \multicolumn{3}{c}{Out of Domain}\\ 
 & Calib.  Accu &  AUROC  &  Cov@Acc=80\% & Calib.  Accu & AUROC  &  Cov@Acc=80\%\\
   \hline
 & \multicolumn{3}{c|}{SQuAD1.1 + HotpotQA}& \multicolumn{3}{c}{SQuAD1.1 + HotpotQA / Other MRQA datasets}\\
 \hline
 MaxProb & 58.2$\pm$0.2 & 58.0$\pm$0.3 & 38.4\% & 56.8$\pm$0.2 & 56.5$\pm$0.2 & 38.3\%\\
 \citet{kamath2020selective} & 62.6$\pm$0.5 & 62.3$\pm$0.7 & 40.9\% &61.2$\pm$0.4 & 60.7$\pm$0.5 & 39.7\%\\
 Ours & 65.8$\pm$0.3 & 66.8$\pm$0.4 & 43.1\% & 63.7$\pm$0.3 & 64.1$\pm$0.3 & 41.6\% \\
%   + CLS  & 65.9$\pm$0.4 & 66.8$\pm$0.5 & 52.9\% & 62.8$\pm$0.3 & 65.2$\pm$0.4 & 51.0\%\\ 
 +  features & 67.4$\pm$0.5 & 68.5$\pm$0.4 & 43.3\% & 65.4$\pm$0.3 & 66.9$\pm$0.3 & 42.7\%\\ 
 +  features + $\context$ & \bf{69.2$\pm$0.4} & \bf{70.3$\pm$0.4}& \bf{44.3\%} & \bf{67.6$\pm$0.4} & \bf{68.8$\pm$0.5}&  \bf{43.9\%} \\
 + features + $\question$ & 66.8$\pm$0.3 & 67.9$\pm$0.3 & 42.4\% & 64.7$\pm$0.4 & 66.2$\pm$0.3 & 42.5\%\\ 
   \hline
 & \multicolumn{3}{c|}{SQuAD1.1 + NQ}& \multicolumn{3}{c}{SQuAD1.1 + NQ / Other MRQA datasets}\\
 \hline
 MaxProb & 64.8$\pm$0.3 & 71.5$\pm$0.3 & 49.2\% & 61.4$\pm$0.2 & 66.7$\pm$0.3 & 45.9\% \\
%  \citep{kamath2020selective}$^*$ & 74.7 & 83.8 & 74.7 & 83.8 \\
 \citet{kamath2020selective} & 68.5$\pm$0.4 & 75.5$\pm$0.6 & 53.4\% & 64.1$\pm$0.6 & 69.2$\pm$0.5 & 51.5\% \\
%  \citep{kamath2020selective} + Ours & $\pm$ & $\pm$ & $\pm$ & $\pm$\\
% + CLS  & 69.1$\pm$0.5 & 76.3$\pm$0.5 & 60.3\% & 62.8$\pm$0.4 & 69.8$\pm$0.6 & 54.5\%\\ 
Ours & 69.5$\pm$0.3 & 76.3$\pm$0.5 & 57.8\% & 64.3$\pm$0.4 & 69.4$\pm$0.4 & 54.3\% \\
 + features & 70.3$\pm$0.4 & 77.0$\pm$0.3 & 59.1\% & 64.9$\pm$0.5 & 70.4$\pm$0.5 & 56.5\%\\
+  features + $\context$  & \bf{73.2$\pm$0.4} & \bf{79.4$\pm$0.3} & \bf{60.7\%} & \bf{66.7$\pm$0.5} & \bf{72.1$\pm$0.5} & \bf{57.6\% }\\
 + features + $\question$ & 72.5$\pm$0.4 & 78.7$\pm$0.3 & 59.3\% & 65.8$\pm$0.5 & 71.4$\pm$0.5& 55.9\% \\
 \bottomrule
  \end{tabular}
 \caption{Calibration results on standard reading comprehension datasets. In the out of domain setting, we first list the training dataset of calibrator, then the test dataset. }
    \label{tab:out_of_domain}
\end{table*}

\paragraph{Ours} represents a calibrator that is trained with the question context embedding, $\phi(q_i, c_i)$ in Eqn \eqref{eq:feature}.  `+ features' refers to augmenting features from \citep{kamath2020selective}, described above. Augmenting the feature sets with question context embeddings from backtranslated questions is denoted as `+$\question$', and augmenting the feature sets with question context embeddings from backtranslated contexts is denoted as `+$\context$' from Eqn. \eqref{eq:bt}.

% the reading comprehension model takes the top $100$ retrieved passages as input and trains a reader model, which optimizes a softmax cross-entropy loss to select the correct passage out of all candidates, and the answer span prediction loss (softmax cross-entropy loss at the start and end indices). The model then selects the answer span with the highest answer span score (sum of the start and end logit score) from the passage with the highest passage score. 

\subsubsection{Open Retrieval QA}
We consider separate calibrators for two different approaches~\cite{karpukhin2020dense,Raffel2020ExploringTL}. 
\paragraph{Extractive (Retrieve-And-Predict)} 
We consider two baseline calibrators: one takes the product of normalized passage score (normalized across all passage candidates) and answer score (normalized across the top 10 answer spans for each passage), and another takes the product of unnormalized passage and answer scores. 

Then, we introduce calibrator augmented with our input example embedding. We include two example embeddings as features: one is the question context embedding as used in the reading comprehension setting (from Eqn~\ref{eq:feature}), and another is the average of the answer span start token representation and the answer span end token representation. 

\paragraph{Generation based (Seq2Seq)} For seq2seq models~\cite{Raffel2020ExploringTL}, the output answer space includes all sentences that can be generated with conditional language model. Thus, instead of MaxProb, we use the likelihood of the generated answer (i.e., the product of the conditional probabilities for each token in the generated answer) as a baseline. Then, we introduce calibrator with our input example embedding (from Eqn~\ref{eq:feature}).  %This roughly corresponds to the MaxProb definition from \citep{hendrycks2016baseline}, which measures the model's score on the top answer candidate. 

%% file: results-5.tex
\section{Results}

\begin{table*}\vspace{-5pt}
 \footnotesize
\centering
 \begin{tabular}{lccc|ccc}
 \toprule
 & \multicolumn{3}{c|}{In Domain} & \multicolumn{3}{c}{Out of Domain}\\ 
 & Calib.  Accu &  AUROC  &  Cov@Acc=80\% & Calib.  Accu & AUROC  &  Cov@Acc=80\%\\ 
 \hline
 & \multicolumn{3}{c|}{SQuAD1.1 Adversarial} &\multicolumn{3}{c}{MRQA / SQuAD1.1 Adversarial}  \\ 
 \hline
 \citet{kamath2020selective}& 52.4$\pm$0.2 & 53.7$\pm$0.4 & 25.4\% & 52.4$\pm$0.2 & 52.2$\pm$0.4 & 24.7\%\\
Ours & 61.1$\pm$0.4 & 63.2$\pm$0.3 & 35.6\% & 53.2$\pm$0.4 & 53.6$\pm$0.3 & 25.3\%\\ 
 + features & 61.4$\pm$0.6 & 63.5$\pm$0.3 & 35.8\% & 53.8$\pm$0.4 & 54.3$\pm$0.4 & 26.8\%\\ 
 + features + $\context$ & \bf{62.8$\pm$0.3} & \bf{65.2$\pm$0.2}& \bf{37.3\%} & \bf{54.9$\pm$0.5} & \bf{55.1$\pm$0.3}& \bf{27.5\%}\\
 + features + $\question$ & 61.6$\pm$0.3 & 63.7$\pm$0.4 & 35.5\% & 53.6$\pm$0.5 & 53.9$\pm$0.5 & 26.6\% \\
 \midrule
  &  \multicolumn{3}{c|}{SQuAD2.0 } &  \multicolumn{3}{c}{MRQA / SQuAD2.0 }\\ 
 \hline
 \citet{kamath2020selective} & 57.6$\pm$0.3 & 59.2$\pm$0.4 & 31.7\% & 54.8$\pm$0.4 & 56.5$\pm$0.5 & 29.6\%\\
 Ours & 58.9$\pm$0.2 & 61.1$\pm$0.2 & 33.8\% & 55.7$\pm$0.3 & 57.4$\pm$0.4 & 30.7\% \\
 + features & 60.1$\pm$0.2 & 61.9$\pm$0.3 & 34.2\% & 56.6$\pm$0.4 & 58.3$\pm$0.5 & 31.6\%\\ 
 + features + $\context$ & 60.2$\pm$0.3 & 61.8$\pm$0.3 & 34.1\% & 56.4$\pm$0.5 & 57.9$\pm$0.4 & 31.2\%\\
 + features + $\question$ & \bf{62.6$\pm$0.4} & \bf{64.3$\pm$0.3} & \bf{35.9\%} & \bf{58.1$\pm$0.4} & \bf{60.4$\pm$0.4} & \bf{32.9\%} \\
 \bottomrule
  \end{tabular}
 \caption{Calibration results on adversarial and unanswerable SQuAD datasets. In the out of domain setting, we first list the training dataset of calibrator, then the test dataset. 
 }
    \label{tab:harder_mc}
\end{table*}

\begin{table*}[h]
  
\footnotesize
\centering
 \begin{tabular}{lclccc}
 \toprule
Model & Answer Acc &Calibrator  & Calib. Accu  & Calib. AUROC  & Cov@Acc=80\%\\
 \midrule

%  DPR & 41.0 &{\citet{kamath2020selective}} & 63.8$\pm$0.3 & 63.4$\pm$0.2 & 5.8\%\\% 9.7$\pm$0.2 & 20.6$\pm$0.2  \\
Extractive (DPR) & 41.0 & Unnormalized scores & 65.9$\pm$0.2 & 65.2$\pm$0.2 & 10.4\%\\%10.3$\pm$0.2 & 23.1$\pm$0.3 \\
 \tiny{\citep{karpukhin2020dense}}&  & Normalized scores & 72.2$\pm$0.4 & 74.5$\pm$0.3 & 28.9\%\\%41.2$\pm$0.1 & 58.6$\pm$0.1\\
 & & Ours (+ Normalized Scores)& \bf{77.3$\pm$0.3} & \bf{78.7$\pm$0.2} &  \bf{30.5\%} \\\midrule
 Generation (T5)& 25.5 &Likelihood & 89.3$\pm$0.1 & 86.6$\pm$0.1 & 10.4\%\\
 \tiny{\citep{Raffel2020ExploringTL}}& & Ours (+ Likelihood) & \bf{91.6$\pm$0.3} & \bf{92.9$\pm$0.1}&  \bf{ 11.3\%}\\ %\bf{41.4$\pm$0.1} & \bf{59.0$\pm$0.1} \\
%  + Back-translation Question& \\
 \bottomrule
  \end{tabular}
 \caption{Calibration results on NQ open retrieval test set for different base QA models and calibration features.}
    \label{tab:opendomain_qa}
\end{table*}
\paragraph{Calibration}
Table~\ref{tab:out_of_domain} reports calibration results on standard reading comprehension datasets. 
The top block displays the performance of calibrators trained on the SQuAD and HotpotQA datasets, 
and the bottom block shows the results of calibrators trained on the SQuAD and NQ datasets. In both settings, the our input example embedding works better than the manual feature set. However, two approaches are complementary in all settings. Interestingly, paraphrasing questions shows gains in Natural Questions but not in other datasets. We hypothesize that organically collected search queries contain more ambiguous and ill-defined queries than crowdsourced questions where questions were based directly on the context. Adding paraphrased context embeddings, on the other hand, shows a modest gain across all settings. Unlike QA models have access to millions of parameters, calibrators, even with our feature set, are provided with very limited information. We hypothesize that augmenting the feature set with paraphrased context enabled the calibrator to gain more information about the example, facilitating higher performance.

Table~\ref{tab:harder_mc} shows the results in more challenging settings: one with adversarial attacks and another containing unanswerable questions. In both settings, we observe sizable gains (5-10\% increase in calibration accuracy) for the in domain setting, but the gains are smaller in out of domain settings. Similar to the Natural Questions dataset, in SQuAD 2.0, which includes adversarially designed questions without an answer, paraphrasing the question is more helpful than paraphrasing the context. On the other hand, in the adversarial setting where contexts are manipulated, paraphrasing contexts is more effective. Overall, our new feature vector shows consistent gain across all datasets and settings.

We present the calibration in open retrieval QA in Table~\ref{tab:opendomain_qa}. Overall, calibrator accuracy is higher compared to RC, partially because the answer accuracy is substantially lower. For example, with generation based model (T5)'s answer accuracy of 25.5, simply predicting incorrectly for every example will give 74.5 calibration accuracy. In both models, internal confidence scores (Likelihood and Normalized scores) provide reasonable calibrator performance, yet adding our feature set improves the performance. In particular, our calibrator shows a larger gain in the DPR setting. Encouraged by this result, we test our calibrator as an answer candidate reranker for top answer candidates from DPR. Despite high calibration accuracy of generation based approach, selective QA performance (Cov@Acc=80\%) is higher with the extractive approach, suggesting comparing calibration performance across models of different accuracy is challenging.

\paragraph{Answer Reranking}
Table~\ref{tab:reranker} shows the results of our calibrator as an answer candidate reranker. The calibrator considers the top 1,000 answer candidates (100 retrieved passages, each with top 10 answer spans) and outputs top candidates based on the calibrator score instead of the model score. We show negligible gains in top 1 accuracy but bigger gains in top 5 accuracy. These small but noticeable gains show potential for using calibrators to improve open retrieval QA performances, where multiple answer candidates are considered.

\begin{table}
 
\centering
\resizebox{0.95\columnwidth}{!}{
 \begin{tabular}{lcc}
 \toprule
  & Top 1 EM & Top 5 EM  \\
 \hline
 DPR  &  41.0 &57.8 \\
% {\citet{kamath2020selective}}  &9.7$\pm$0.2 & 20.6$\pm$0.2  \\
Unnormalized scores &   10.3$\pm$0.2 & 23.1$\pm$0.3 \\
Normalized scores &  41.2$\pm$0.1 & 58.6$\pm$0.1\\
 Ours (+ Normalized scores) &   \bf{41.4$\pm$0.1} & \bf{59.0$\pm$0.1} \\
%  + Back-translation Question& \\
 \bottomrule
  \end{tabular}}
 \caption{Results on open domain question answering in NQ. The calibrator is used as a reranker for selecting the top answer span out of 1,000 answer spans (10 answer spans per each of 100 retrieved passages). }
    \label{tab:reranker}
\end{table}

%% file: analysis-6.tex
\section{Analysis}
%We provide analysis on different components of our calibrator. %First, we examine our new question context embedding, whether it is important to use QA model or general language model can replace it. Next, we investigate the quality of our data augmentation through back translation. %Lastly, we present a visu
\begin{table}
 \small
\centering
 \begin{tabular}{lc|c}
 \toprule
  &{In Domain} & {Out of Domain}\\ 
% & Calib.  Accu &  AUROC  &  Cov@Acc=80\% & Calib.  Accu & AUROC  &  Cov@Acc=80\%\\
   \hline
 & {SQuAD1.1 + Hotpot QA}& {Other MRQA datasets}\\
 \midrule
    CLS  & 63.5$\pm$0.4  & 62.3$\pm$0.5 \\
%   + CLS  & 65.9$\pm$0.4 & 66.8$\pm$0.5 & -\% & 62.8$\pm$0.3 & 65.2$\pm$0.4 & -\%\\ & 65.2$\pm$0.4 & 41.8\%& 62.6$\pm$0.3 & 40.3\%

Ours & 65.8$\pm$0.3  & 63.7$\pm$0.3 \\%& 66.8$\pm$0.4 & 43.1\% & 64.1$\pm$0.3 & 41.6\%
 
%  + Final Layer Hidden & 67.4$\pm$0.5 & 68.5$\pm$0.4 & -\% & 65.4$\pm$0.3 & 66.9$\pm$0.3 & -\%\\
  Diff. & 2.3\% &1.4\%\\ % 1.6 & 1.3\% & 
 \bottomrule
  \end{tabular}
 \caption{CLS token ablation results, all numbers refer to calibration accuracy. Using CLS token as a feature shows a strong calibration performance, lagging behind question context encoding from the RC model only by a few points. The gap is even smaller in out of domain setting. }
    \label{tab:cls_token_ablation}
\end{table}

\subsection{Task-Agnostic Representation vs. Representation from QA Model}\label{subsec:task-agnostic}
Our study has shown that input example embedding is very useful, adding complementary power to model confidence features. Based on this result, we further ask the question, is it possible to build a calibrator without accessing the model parameters, but only a small amount of calibration training data (which consists of questions, context, and whether the model's prediction is correct or incorrect)? We train a calibrator that does not have any access to the QA model parameters and only takes the model's predictions on a small set of training data (a couple thousand of QA examples). This calibrator uses a standard pretrained language model (BERT) to encode [CLS$; (q_i, c_i)$] and takes the final layer hidden representation of the [CLS] token as a feature. Table~\ref{tab:cls_token_ablation} shows the performance of the [CLS] token classifier. 
Surprisingly, this calibrator outperforms the MaxProb baseline (in Table~\ref{tab:out_of_domain}) in all settings and outperforms ~\citet{kamath2020selective} (in Table~\ref{tab:out_of_domain}) in most settings, indicating information about the question and context might be more useful than the QA model's confidence. Using the input example embedding from the QA model shows only 1-3 point gains than using the CLS token embedding. This trend holds for across various settings (more results in Table~\ref{tab:cls_token_ablation_full} in Appendix). 

\subsection{Quality of Back Translation}\label{subsec:qual_backtranslation}
Question paraphrasing~\cite{dong-etal-2017-learning} can improve performances of QA models. Similarly, both question and context paraphrasing improves calibration performance. In this section, we investigate the quality of backtranslation used in our study. We manually inspect 100 question paraphrasing from SQuAD 2.0 dataset. 71 examples maintain the original meaning, 12 examples change its meanings, and 17 examples are hard to distinguish.
 One common pattern for meaning change is when proper nouns in the original sentences are missing and incorrectly translated (e.g. John Calvin $\rightarrow$ Jean Calvin).
%Quantitatively, in the 100 examples,
\begin{table}
    \centering
    \footnotesize
    \begin{tabular}{c|p{21em}}
         \hline
         $q$ & In what country is Normandy located? \\
         $q'$ & What country is Normandy in? \\
         \hline
         $q$ & When did Edward return? \\
         $q'$  &When did Edward come back? \\
         \hline
         $q$ & How would one write T(n) = 7n2 + 15n + 40 in big O notation?\\
         $q'$  &How do you write T(n) = 7n2 + 15n + 40? \\
         \hline
         $q$ & What kind of arches does Norman architecture have? \\
         $q'$  & What kind of arches does Norman's building have? \\
         \hline
    \end{tabular}
    \caption{Question back translation samples from SQuAD 2.0 dataset. The first row (q) refers to the original question, and the second row ($q'$) refers to backtranslated question. In the third example, back translation introduces an error. }
    \label{tab:my_label}
\end{table}

We study how much variability is introduced during paraphrasing by studying divergence between the original sentence and the paraphrased sentence. 
We calculate the sentence BLEU score with \emph{NLTK} \cite{bird2009nltk}, using the original text as source and the back-translated text as target for both question paraphrasing and context paraphrasing. The average sentence BLEU score is larger than 0.55 for all datasets, indicating back-translation introduces relatively minor changes in phrasing.

\begin{figure}
\centering
\footnotesize
\vspace{-1em}
\includegraphics[width=0.45\textwidth]{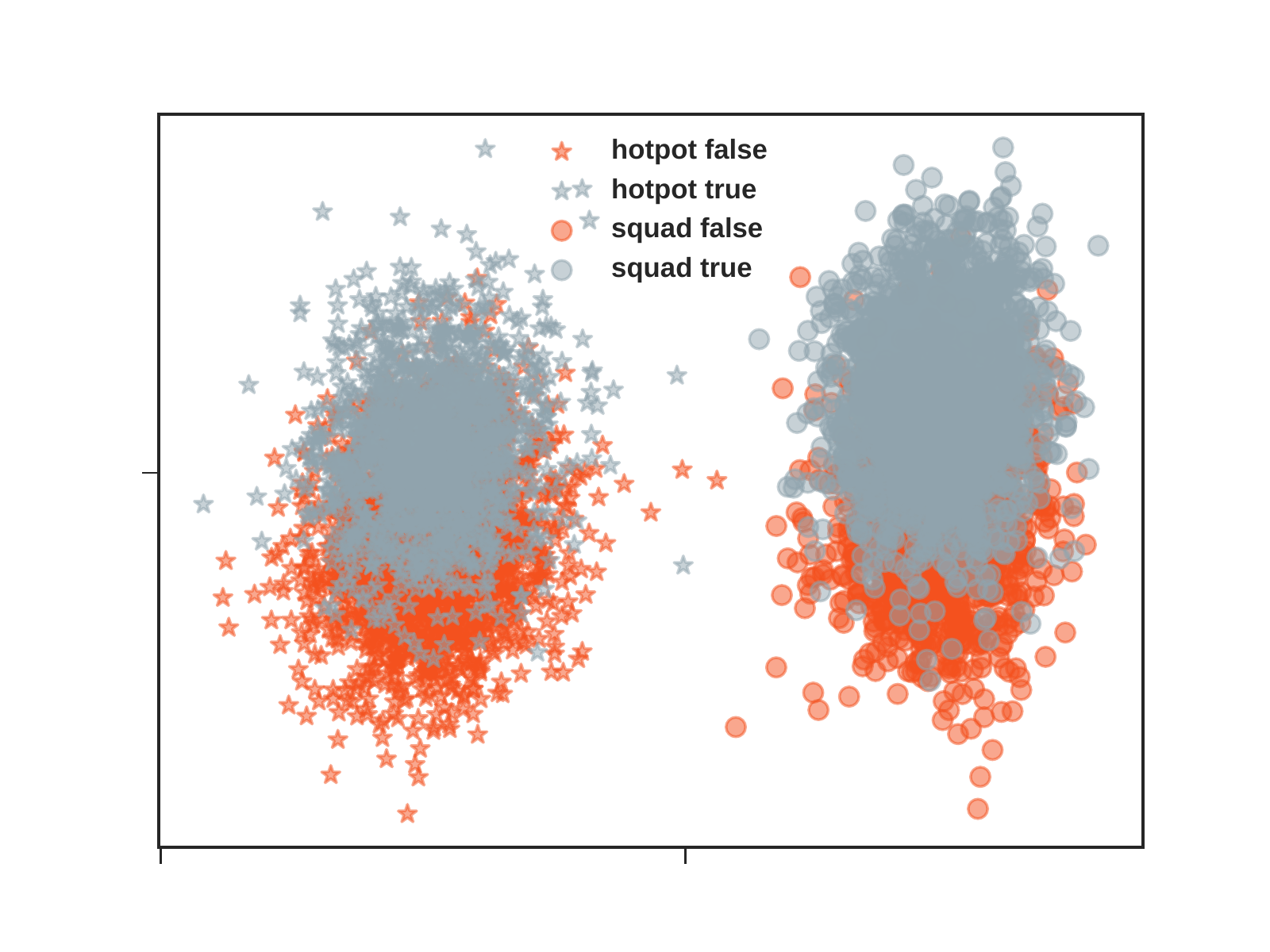}
\vspace{-1.3em}
\caption{A visualization for the input example embedding from HotPotQA and SQuAD datasets. 
We denote the data domain by markers with different shapes and denote the correctness with different colors.
% the label distinguish the domain and whether the example is correct, 
The X-axis and Y-axis denote the first and second dimensions extracted by linear discriminant analysis, respectively.}\vspace{-0.4em}
\label{fig:hidden_represetation}
\end{figure}

\paragraph{Visualization}
Figure~\ref{fig:hidden_represetation} shows a visualization of the question context embeddings from HotpotQA and SQuAD. 
We use linear discriminant analysis~\citep{pedregosa2011scikit} to plot input example embeddings and observe that embeddings from the same dataset are closer to each other. It demonstrates that embeddings are almost linearly separable between domains, but it is much harder to distinguish correct answers from incorrect ones.

\begin{table}
 \footnotesize
\centering
 \begin{tabular}{lc|c}
 \toprule
  & {In Domain} & {Out of Domain}\\ 
 & {SQuAD1.1 + NQ}& { Other MRQA datasts}\\
 \hline
Xgboost & 67.4$\pm$0.5  & 65.4$\pm$0.3  \\  
LR  & 66.6$\pm$0.3  & 64.7$\pm$0.3 \\ 
KNN & 66.3$\pm$0.2  & 64.6$\pm$0.3 \\
 \bottomrule
  \end{tabular}
 \caption{Ablation study on different classifiers with features (Ours + features). All numbers refer to calibration accuracy.}
    \label{tab:ablation_classifier}
\end{table}

\paragraph{Choice of Calibrator Architecture} 
We test if our results are sensitive to the choice of classifiers: XGBoost, logistic regression (LR), and k-nearest neighbors (KNN). 
Table~\ref{tab:ablation_classifier} indicates our gains hold across different classifiers. Full experimental results can be found in Appendix.
%For example, for reading comprehension out of domain setting where the calibrator is trained on SQuAD 1.1 + Hotpot, the  Cov@Acc=80\% for xgboost, logistic regression and KNN are 42.7\%, 42.3\%, and 41.8\% respectively.
%The difference is slight.

% \vspace{-10pt}

\paragraph{Choice of Pivot Language}
We test whether the choice of pivot language in backtranslation impacts performances. We find little difference between pivoting through German or French (See Table~\ref{tab:backtrans_ablation}).

\begin{table}[t]
 \small
\centering
\resizebox{1.0\columnwidth}{!}{
 \begin{tabular}{lc|c|c}
 \toprule
  &Lang-& \multicolumn{1}{c|}{In Domain} & \multicolumn{1}{c}{Out of Domain}\\ 

%  & \multicolumn{3}{c|}{\bf{SQuAD1.1 + Hotpot}}& \multicolumn{3}{c}{\bf{Other}}\\
%  \hline
 &  uage   &\multicolumn{1}{c|}{{SQuAD1.1 + NQ}}& \multicolumn{1}{c}{{Other MRQA datasets }}\\
 \hline
%  Back-translation Context (Fr) & 73.2$\pm$0.4 & 79.4$\pm$0.3 & 63.5\% & 66.7$\pm$0.5 & 72.1$\pm$0.5 & 57.3\% \\
 $\context$ &FR & 73.2$\pm$0.4 &  66.7$\pm$0.5\\% & 72.1$\pm$0.5  \\
 
  $\context$ &DE & 73.0$\pm$0.4 &  66.5$\pm$0.5 \\%& 71.9$\pm$0.6 \\
 \hline
 $\question$ &FR& 72.5$\pm$0.4  & 65.8$\pm$0.5 \\%& 71.4$\pm$0.5 \\
$\question$ &DE & 72.8$\pm$0.3 & 66.1$\pm$0.4 \\%& 71.7$\pm$0.5  \\
 \bottomrule
  \end{tabular}}
 \caption{Calibration accuracy for different pivot languages: French vs. German, using calibrator with features (Ours + features).
 }
    \label{tab:backtrans_ablation}
\end{table}

%% file: related-7.tex
\section{Related Work}
\paragraph{Calibration in NLP}
Calibration has become an important topic in NLP as well as general machine learning~\cite{Guo2017OnCO, pleiss2017fairness, fan2021contextual} as confidence scores from calibrators can be useful for the error correction process~\cite{feng2004using}. Calibration has been studied in natural language inference, commonsense reasoning~\cite{Desai2020CalibrationOP,Varshney2020ItsBT}, dialogue systems ~\citep{Mielke2020LinguisticCT}, semantic parsing~\cite{Dong2018ConfidenceMF}, coreference resolution~\cite{Nguyen2015PosteriorCA} and sequence labeling~\citep{Jagannatha2020CalibratingSO}. 

In question answering, ~\citet{kamath2020selective}'s study on selective question answering inspired our work. We measure the calibration performance with calibrator accuracy, AUROC, and coverage at accuracy. Expected Calibration Error (ECE) \cite{guo2017calibration} is another commonly used metric for calibration performance, but we consider calibrator as a binary classifier at here.
~\citet{Jagannatha2020CalibratingSO} also studies calibration in reading comprehension, using language model perplexity and model's confidence as features. Language model perplexity coarsely and indirectly captures information about the question and context. We propose an improved feature space and thoroughly test it in challenging settings, e.g., adversarial RC, unanswerable RC, and open retrieval QA.

\paragraph{Calibration During Training} Recent work in QA introduces an answer verification step \citep{tan2018know, hu2019read+, wang2020no} at the end of the pipeline. During the training, this verifier module takes the questions, answers, or MRC model's state as inputs and determines the answers' validity. Then, the validity score is used to update the model parameters during training. Thus, the validator is jointly trained with the MRC model. While this is conceptually similar to our set up, instead of tying the calibrator into the model, we design a universal post-hoc calibrator that can be easily applied to any model architecture.

\paragraph{Calibration with Ensembles}
Ensemble diversity has been used to improve uncertainty estimation and calibration \citep[e.g.][]{Raftery2005UsingBM, Stickland2020DiverseEI}. While it is effective, calibration with model ensembling is usually expensive and time consuming~\cite{Zhou2002EnsemblingNN,Zhou2018DiverseEE}. Our calibrator is an offline postprocessing step that does not require further training of the original model.

%% file: conclusion.tex
\section{Conclusion}
We introduce a richer feature space for question answering calibrators with question and context embeddings and paraphrase-augmented inputs. Our work suggests deciding the correctness of a QA system depends on both the semantics of the question-context and the confidence of the model. We thoroughly test our calibrator in domain shift, adversarial, and open domain QA settings. The experiments show noticeable gains in performance across all settings.
We further demonstrate our calibrator's general applicability by using it as a reranker in extractive open domain QA. To summarize, our calibrator is simple, effective and general, with potential to be incorporated into
existing models or extended for other NLP tasks.
% diverse model architectures or other NLP tasks.% as a post hoc processing step. 

%only studies a framework to evaluate how models can judiciously abstain in these challenging environments but also can be easier to find difficult questions. 

%% file: appendix.tex
\section*{Appendix}
\appendix\section{Additional Experimental Results}
% \begin{table}[h]
%  \footnotesize
% \centering
%  \begin{tabular}{lccc|ccc}
%  \toprule
%   & \multicolumn{3}{c|}{In Domain} & \multicolumn{3}{c}{Out of Domain}\\ 
%  & Calib.  Accu &  AUROC  &  Cov@Acc=80\% & Calib.  Accu & AUROC  &  Cov@Acc=80\%\\
%  & \multicolumn{3}{c|}{SQuAD1.1 + NQ}& \multicolumn{3}{c}{SQuAD1.1 + NQ / Other MRQA datasets}\\
%  \hline
% %  \citep{kamath2020selective}$^*$ & 74.7 & 83.8 & 74.7 & 83.8 \\
% %  \citep{kamath2020selective} + Ours & $\pm$ & $\pm$ & $\pm$ & $\pm$\\
% CLS  & 66.8$\pm$0.3 & 74.0$\pm$0.5 & 58.5\% & 62.8$\pm$0.4 & 67.8$\pm$0.4 & 57.6\%\\ 
% Ours & 69.5$\pm$0.3 & 76.3$\pm$0.5 & 62.8\% & 64.3$\pm$0.4 & 69.4$\pm$0.4 & 59.3\% \\
% % + CLS  & 69.1$\pm$0.5 & 76.3$\pm$0.5 & -\% & 62.8$\pm$0.4 & 69.8$\pm$0.6 & -\%\\ 
% %  + Final Layer Hidden & 70.3$\pm$0.4 & 77.0$\pm$0.3 & -\% & 64.9$\pm$0.5 & 70.4$\pm$0.5 & -\%\\
%  Difference & 2.7 & 2.3 & 4.3\% & 1.5& 1.6 & 1.7\%\\ 
%  \bottomrule
%   \end{tabular}
%  \caption{CLS token ablation results on reading comprehension.}
%     \label{tab:cls_token_ablation_full}
% \end{table}
\begin{table*}[h]
 \footnotesize
\centering
 \begin{tabular}{lccc|ccc}
 \toprule
  & \multicolumn{3}{c|}{In Domain} & \multicolumn{3}{c}{Out of Domain}\\ 
 & Calib.  Accu &  AUROC  &  Cov@Acc=80\% & Calib.  Accu & AUROC  &  Cov@Acc=80\%\\
  \hline
 & \multicolumn{3}{c|}{SQuAD1.1 + Hotpot}& \multicolumn{3}{c}{SQuAD1.1 + HotpotQA / Other MRQA datasets}\\
 \hline
    CLS  & 63.5$\pm$0.4 & 65.2$\pm$0.4 & 41.8\% & 62.3$\pm$0.5 & 62.6$\pm$0.3 & 40.3\%\\
%   + CLS  & 65.9$\pm$0.4 & 66.8$\pm$0.5 & -\% & 62.8$\pm$0.3 & 65.2$\pm$0.4 & -\%\\ 

Ours & 65.8$\pm$0.3 & 66.8$\pm$0.4 & 43.1\% & 63.7$\pm$0.3 & 64.1$\pm$0.3 & 41.6\% \\
 
%  + Final Layer Hidden & 67.4$\pm$0.5 & 68.5$\pm$0.4 & -\% & 65.4$\pm$0.3 & 66.9$\pm$0.3 & -\%\\
  Difference & 2.3 & 1.6 & 1.3\% & 1.4& 1.5 & 1.3\%\\ 
  \hline
 & \multicolumn{3}{c|}{SQuAD1.1 + NQ}& \multicolumn{3}{c}{SQuAD1.1 + NQ / Other MRQA datasets}\\
 \hline
%  \citep{kamath2020selective}$^*$ & 74.7 & 83.8 & 74.7 & 83.8 \\
%  \citep{kamath2020selective} + Ours & $\pm$ & $\pm$ & $\pm$ & $\pm$\\
CLS  & 66.8$\pm$0.3 & 74.0$\pm$0.5 & 58.5\% & 62.8$\pm$0.4 & 67.8$\pm$0.4 & 57.6\%\\ 
Ours & 69.5$\pm$0.3 & 76.3$\pm$0.5 & 62.8\% & 64.3$\pm$0.4 & 69.4$\pm$0.4 & 59.3\% \\
% + CLS  & 69.1$\pm$0.5 & 76.3$\pm$0.5 & -\% & 62.8$\pm$0.4 & 69.8$\pm$0.6 & -\%\\ 
%  + Final Layer Hidden & 70.3$\pm$0.4 & 77.0$\pm$0.3 & -\% & 64.9$\pm$0.5 & 70.4$\pm$0.5 & -\%\\
 Difference & 2.7 & 2.3 & 4.3\% & 1.5& 1.6 & 1.7\%\\ 
 \bottomrule
  \end{tabular}
 \caption{CLS token ablation results on reading comprehension.}
    \label{tab:cls_token_ablation_full}
\end{table*}

\begin{table}[h]
 \footnotesize
\centering
 \begin{tabular}{lccc|ccc}
 \toprule
  & \multicolumn{3}{c|}{In Domain} & \multicolumn{3}{c}{Out of Domain}\\ 
 & Calib.  Accu &  AUROC  &  Cov@Acc=80\% & Calib.  Accu & AUROC  &  Cov@Acc=80\%\\
   \hline
 & \multicolumn{3}{c|}{SQuAD1.1 + Hotpot}& \multicolumn{3}{c}{SQuAD1.1 + HotpotQA / Other MRQA datasets}\\
 \hline
Xgboost & 67.4$\pm$0.5 & 68.5$\pm$0.4 & 43.3\% & 65.4$\pm$0.3 & 66.9$\pm$0.3 & 42.7\% \\  
Logistic Regression  & 66.6$\pm$0.3 & 67.3$\pm$0.3 & 42.6\% & 64.7$\pm$0.3 & 66.1$\pm$0.3 & 42.3\% \\ 
KNN & 66.3$\pm$0.2 & 67.0$\pm$0.3 & 42.1\% & 64.6$\pm$0.3 & 65.8$\pm$0.2 & 41.8\% \\
   \hline
 & \multicolumn{3}{c|}{SQuAD1.1 + NQ}& \multicolumn{3}{c}{SQuAD1.1 + NQ / Other MRQA datasets}\\
 \hline
Xgboost & 70.3$\pm$0.4 & 77.0$\pm$0.3 & 59.1\% & 64.9$\pm$0.5 & 70.4$\pm$0.5 & 56.5\%\\ 
Logistic Regression & 69.7$\pm$0.3 & 76.3$\pm$0.2 & 58.6\% & 64.2$\pm$0.4 & 69.7$\pm$0.4 & 56.1\%\\ 
KNN & 68.9$\pm$0.2 & 75.8$\pm$0.2 & 58.3\% & 63.8$\pm$0.3 & 69.3$\pm$0.3 & 55.6\%\\ 
 \bottomrule
  \end{tabular}
 \caption{Ablation study on different classifiers with features (Ours + features).
 }
    \label{tab:ablation_classifier_full}
\end{table}

\section{Hyperparameters and Training Details}

A binary classifier is trained using the gradient boosting library XGBoost~\cite{chen2016xgboost}. We finetune the following hyper-parameters, colsample by level, colsample by node, colsample by tree, learning rate, and the number of estimators on the development set.
We use the following search space: colsample by level/mode/tree is set to the same  value and selected from \{0.0, 0.1, 0.2, 0.3, 0.4, 0.5\}, the learning rate and number of estimators are selected from \{0.01, 0.1, 0.2, 0.5\} and \{5, 25, 50, 100\}, respectively.
These hyper-parameters are chosen based on  the performance on the validation set.

For base QA models, we mostly following the hyperparameters used in the original work (e.g., batch size 32 \& learning rate of $5\times10^{-5}$ for BERT-base SQuAD 1.1 model). All calibrators are trained five times, each with different data partitions and random seeds. We report the variances in the results. Our calibrator does not share its weights with the base QA models.